\documentclass[runningheads]{llncs}
\usepackage{comment}

\usepackage{amsmath,amssymb} 
\usepackage{color}

\newcommand{\stylelinewidth}{\linewidth}

\usepackage{epsfig}
\usepackage{graphicx}
\usepackage{amsmath}
\usepackage{amssymb}
\usepackage{booktabs} 
\usepackage{enumitem}
\usepackage{textcomp}
\usepackage{caption}

\usepackage[pagebackref=true,breaklinks=true,letterpaper=true,bookmarks=false]{hyperref}

\usepackage{overpic}
\usepackage{multirow} 
\usepackage{diagbox}
\usepackage[T1]{fontenc} 
\usepackage[style=american]{csquotes} 
\usepackage{mathtools}
\usepackage{ulem}  

\usepackage{color}
\definecolor{turquoise}{cmyk}{0.65,0,0.1,0.3}
\definecolor{purple}{rgb}{0.65,0,0.65}
\definecolor{dark_green}{rgb}{0, 0.5, 0}
\definecolor{orange}{rgb}{0.8, 0.6, 0.2}
\definecolor{red}{rgb}{0.8, 0.2, 0.2}
\definecolor{blueish}{rgb}{0.0, 0.7, 1}
\definecolor{light_gray}{rgb}{0.7, 0.7, .7}
\definecolor{pink}{rgb}{1, 0, 1}
\definecolor{dark_red}{rgb}{0.5, 0, 0}
\definecolor{corn}{rgb}{0.89, 0.82, 0.04}

\newcommand{\hide}[1]{{}} 

\newcommand{\Figure}[1]{Figure~\ref{fig:#1}}

\newcommand{\Table}[1]{Table~\ref{tbl:#1}}
\newcommand{\eq}[1]{(\ref{eq:#1})}

\newcommand{\Equation}[1]{Equation~\ref{eq:#1}}

\newcommand{\Section}[1]{Section~\ref{sec:#1}}

\usepackage{currfile}

\usepackage{blindtext}

\renewcommand{\paragraph}[1]{\vspace{.5em}\noindent\textbf{#1}.~}

\DeclareMathAlphabet\mathbfcal{OMS}{cmsy}{b}{n}

\newcommand{\CIRCLE}[1]{\raisebox{.5pt}{\footnotesize \textcircled{\raisebox{-.6pt}{#1}}}}

\usepackage{pifont}
\newcommand{\cmark}{\ding{51}}%
\newcommand{\xmark}{\ding{55}}%

\usepackage{mathtools}

\newcommand{\V}{\mathbf{V}}

\newcommand{\point}{\mathbf{x}}
\newcommand{\vertex}{\mathbf{v}}
\newcommand{\T}{\mathbf{B}}
\newcommand{\pose}{\boldsymbol{\theta}}
\newcommand{\shape}{\boldsymbol{\beta}}
\newcommand{\poses}{\boldsymbol{\Theta}}

\newcommand{\R}{\mathbb{R}}
\newcommand{\x}{\mathbf{x}}
\newcommand{\object}{\mathcal{O}}

\newcommand{\normal}{\mathcal{N}}
\newcommand{\SDF}{\Phi}

\newcommand{\depth}{\mathbf{D}}
\newcommand{\given}{|}
\newcommand{\expect}[2]{\mathbb{E}_{#1}\left[ #2 \right]}

\newcommand{\rest}{\bar}
\newcommand{\trans}{\mathbf{t}}

\newcommand{\MLP}{\text{MLP}}

\newcommand{\loss}{\mathcal{L}}

\DeclareMathOperator*{\argmax}{arg\,max}
\DeclareMathOperator*{\argmin}{arg\,min}

\begin{document}
\pagestyle{headings}
\mainmatter
\def\ECCVSubNumber{344}  

\title{NASA \\ Neural Articulated Shape Approximation} 


\titlerunning{Neural Articulated Shape Approximation}
%

\author{Boyang Deng\inst{1}
\and 
JP Lewis\inst{1}
\and
Timothy Jeruzalski\inst{1}
\and
Gerard Pons-Moll\inst{2}
\and \\
Geoffrey Hinton\inst{1}
\and
Mohammad Norouzi\inst{1}
\and
Andrea Tagliasacchi\inst{1,3}
} 
\authorrunning{B. Deng et al.}
%
\institute{
Google Research \and MPI for Informatics, Saarland Informatics Campus, Germany \and 
University of Toronto, Canada
} 
\maketitle

\begin{abstract}
Efficient representation of articulated objects such as human bodies is an important problem in computer vision and graphics.
To efficiently simulate deformation, existing approaches represent 3D objects using polygonal meshes and deform them using skinning techniques.
This paper introduces neural articulated shape approximation (NASA), an alternative framework that enables representation of articulated deformable objects using neural indicator functions that are conditioned on pose.
Occupancy testing using NASA is straightforward, circumventing the complexity of meshes and the issue of water-tightness. 
We demonstrate the effectiveness of NASA for 3D tracking applications, and discuss other potential extensions.

\end{abstract}

\begin{figure}[t]
\begin{center}
\begin{overpic} 
[width=\stylelinewidth]
{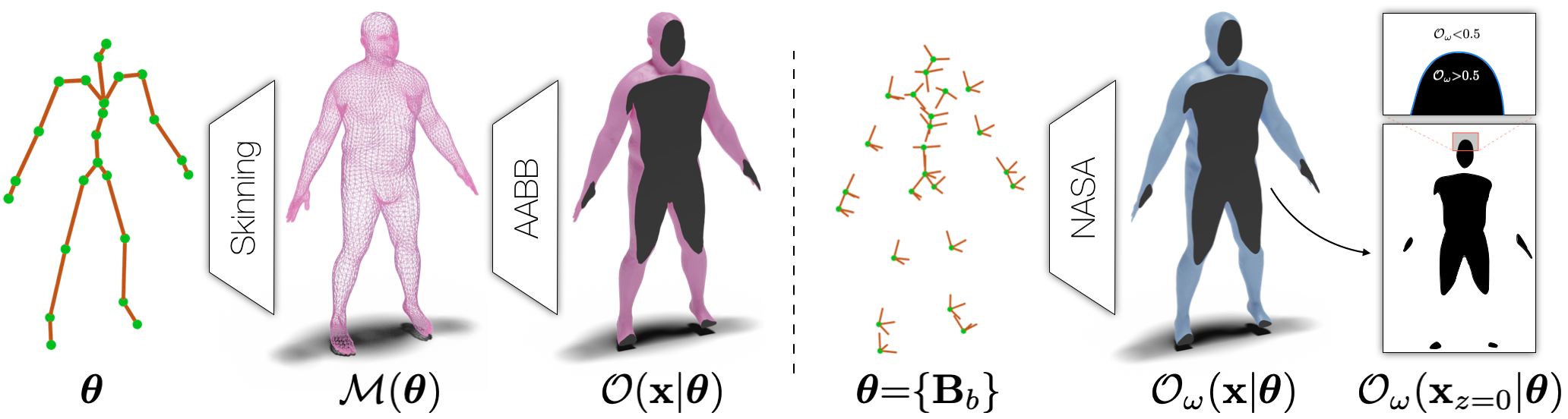}
\end{overpic}
\end{center}
\vspace{-1em}
\caption{
\textbf{Teaser -- }
(left) Traditional articulated models map pose parameters $\pose$ to a polygonal mesh $\mathcal{M}(\pose)$ via linear blend skinning; if one desires to \textit{query} the occupancy of this representation, acceleration data structures need to be computed (e.g. axis align bounding box tree).
(right) Conversely, NASA learns an \textit{implicit} neural occupancy $\object_\omega$, which can be queried directly.
}
\label{fig:teaser}
\end{figure}
\section{Introduction}
There has been a surge of recent interest in representing 3D geometry using implicit functions parameterized by neural networks~\cite{occnet,atlasnet,deepsdf,chibane20ifnet}. Such representations are flexible, continuous, and differentiable.
Neural implicit functions are useful for ``inverse graphics'' pipelines for scene
understanding~\cite{tfgraphics}, as back propagation through differentiable representations of 3D geometry is often required.
That said, neural models of \textit{articulated} objects have received little attention. \linebreak
Articulated objects are particularly important to represent animals and humans, which are central in many applications such as computer games and animated movies, as well as augmented and virtual reality.

Although parametric models of human body such as SMPL~\cite{smpl} have been integrated into neural network frameworks for self-supervision~\cite{kanazawa_cvpr18,omran2018NBF,pavlakos2018humanshape,tung2017self}, these approaches depend heavily on polygonal
mesh representations. 
Mesh representations require expert supervision to construct, and are not flexible for capturing topology variations.
Furthermore, geometric representations often should fulfill several purposes simultaneously such as modeling the surface for rendering, or representing the volume to test intersections with the environment, which are not trivial when polygonal meshes are used~\cite{winding}.
Although neural models have been used in the context of articulated deformation~\cite{sig18skinning}, they \textit{relegate} query execution to classical acceleration data structures, thus sacrificing full differentiability.

Our method represents articulated objects with a neural model, which outputs a differentiable occupancy of the articulated body in a specific pose. 
Like previous geometric learning efforts~\cite{occnet,cvxnet,deepsdf,imnet}, we represent geometry by \textit{indicator functions}~{--}~also referred to as occupancy functions~{--}~that evaluate to $1$ inside the object and $0$ otherwise.
Unlike previous approaches, which focused on collections of static objects described by (unknown) shape parameters, we look at learning indicator functions as we vary \textit{pose parameters}, which will be discovered by training on animation sequences.
We show that existing methods~\cite{occnet,deepsdf,imnet} cannot encode pose variation reliably, because it is hard to learn the occupancy of every point in space as a function of a latent pose vector.

Instead, we introduce NASA, a neural decoder that exploits the structure of the underlying deformation driving the articulated object.
Exploiting the fact that 3D geometry in \textit{local} body part coordinates does not significantly change with pose, we classify the occupancy of 3D points as seen from the coordinate frame of each part.
Our main architecture combines a collection of per-part learnable indicator functions with a per-part pose encoder to model localized non-rigid deformations. 
This leads to a significant boost in generalization to unseen poses,
while retaining the useful properties of existing methods: differentiability, ease of spatial queries such as intersection testing, and continuous surface outputs.
To demonstrate the flexibility of NASA, we use it to track point clouds by finding the maximum likelihood estimate of the pose 
under NASA's occupancy model.
In contrast to mesh based trackers which are complex to implement, 
our tracker requires a few lines of code and is fully differentiable.
Overall, our contributions include:
\begin{enumerate}[leftmargin=*]
\item We propose a neural model of articulated objects to predict differentiable occupancy as a function of pose -- the core idea is to model shapes by networks that encode a piecewise decomposition;
\item The results on learning 3D body deformation outperform previous geometric learning algorithms~\cite{deepsdf,imnet,deepsdf}, and our surface reconstruction accuracy approaches that of mesh-based statistical body models~\cite{smpl};
\item The differentiable occupancy supports constant-time queries ($.06$~ms/query on an NVIDIA GTX~1080), avoiding the need to convert to separate representations, or the dynamic update of spatial acceleration data structures;
\item We derive a technique that employs occupancy functions for tracking 3D geometry via an implicit occupancy template, without the need to ever compute distance functions.
\end{enumerate}
\section{Related work}
Neural articulated shape approximation provides a single framework that addresses problems that have previously been approached separately.
The related literature thus includes a number of works across several different research topics.

\paragraph{Skinning algorithms} 
Efficient articulated deformation is traditionally accomplished with a skinning algorithm that deforms vertices of a mesh surface as the joints of an underlying abstract skeleton change. The classic linear blend skinning~(LBS) algorithm expresses the deformed vertex as a weighted sum of that vertex rigidly transformed by several adjacent bones;~see \cite{skinningcourse} for details.
LBS is widely used in computer games, and is a core ingredient of some popular vision models~\cite{smpl}.
Mesh sequences of general (not necessarily articulated) deforming objects have also been represented with skinning for the purposes of compression and manipulation, using a collection of non-hierarchical ``bones''~(i.e.~transformations) discovered with clustering~\cite{skinningmeshanim,skinningdecomp}.
LBS has well-known disadvantages: the deformation has a simple algorithmic form that cannot produce pose-dependent detail,
it results in characteristic volume-loss effects such as the ``collapsing elbow'' and ``candy wrapper'' artifacts \cite[Figs.~2,3]{posespacedeformation}, and for best results the weights must be \textit{manually} painted by artists.
It is possible to add pose-dependent detail with a shallow or deep net regression~\cite{posespacedeformation,sig18skinning}, but this process operates as a correction to classical LBS deformation.

\paragraph{Object intersection queries}
Registration, template matching, 3D tracking, collision detection, and other tasks require efficient inside/outside tests.
A disadvantage of polygonal meshes is that they do not efficiently support these queries, as meshes often contain thousands of individual triangles that must be tested for each query.
This has led to the development of a variety of spatial data structures to accelerate point-object queries~\cite{Lin96collisiondetection,sametSDSbook}, including voxel grids, octrees, kdtrees, and others.
In the case of deforming objects, the spatial data structure must be repeatedly rebuilt as the object deforms.
A further problem is that typically meshes may be constructed (or deformed) without regard to being ``watertight'' and thus do not have a clearly defined interior~\cite{winding}.

\paragraph{Part-based representations}
For object intersection queries on articulated objects, it can be more efficient to approximate the overall shape in terms of a moving collection of rigid parts, such as spheres or ellipsoids, that support efficient querying~\cite{hadjust}; see
\textbf{supplementary material} for further discussion.
Unfortunately this has the drawback of introducing a second approximate representation that does not exactly match the originally desired deformation.
A further core challenge, and subject of continuing research, is the automatic creation of such \textit{part-based} representations~\cite{anguelovArticulated,hierarchicalsegment,jointshapeseg}.
%
Unsupervised part discovery has been recently tackled by a number of deep learning approaches~\cite{cvxnet,lorenz2019unsupervised,baenet,cerberus,sdmnet}.
In general these methods address analysis and correspondence across shape collections, but do not target accurate representations of articulated objects, and do not account for pose-dependent deformations.

\paragraph{Neural implicit object representation}
Finally, several recent works represent objects with neural implicit functions~\cite{occnet,imnet,deepsdf}.
These works focus on the neural representation of \textit{static} shapes in an \textit{aligned} canonical frame and do not target the modeling of transformations.
Our core contributions are to show that these architectures have difficulties in representing complex and detailed \textit{articulated} objects~(e.g. human bodies), and that a simple architectural change can address these shortcomings.
Comparisons to these closely related works will be revisited in more depth in \Section{evaluation}.
\begin{figure*}[t]
\centering
\begin{overpic} 
[width=\stylelinewidth]
{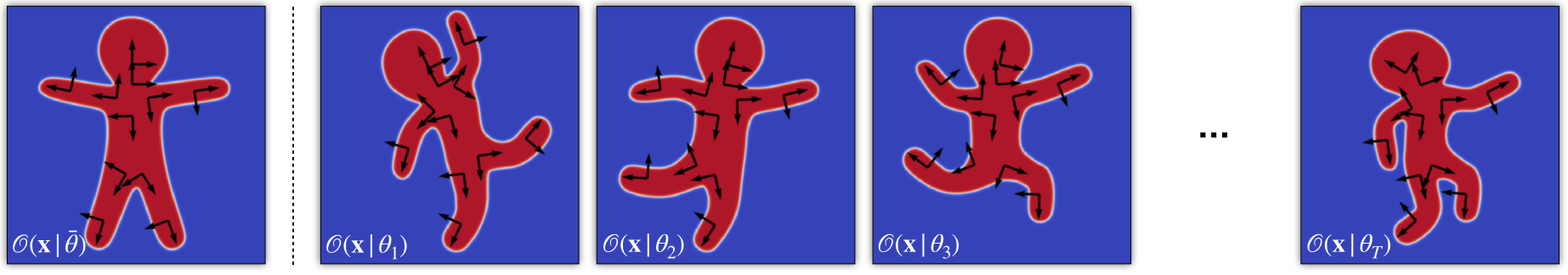}
\end{overpic}
\vspace{-1.5em} 
\caption{
\textbf{Notation} --
(left) The ground truth occupancy $\object(\x \given \bar\pose)$ in the rest frame and the pose parameters $\bar\pose {\equiv} \{\bar\T_b\}_{b=1}^B$ representing the transformations of $B$ bones.
(right) $T$ frames of an animation associated with pose parameters~$\{\pose_t\}_{t=1}^T$ with corresponding occupancy~$\{\object(\x \given \pose_t)\}_{t=1}^T$; each $\pose_t$ encodes the transformations of $B$ bones.
Note we shorthand $\{*_y\}$ to indicate an ordered set $\{*_y\}_{y=1}^Y$.
}
\label{fig:notation}
\end{figure*}
  
\section{Neural Articulated Shape Approximation}
\label{sec:nasa}

This paper investigates the use of neural networks and implicit functions for modeling articulated shapes in $\R^d$.
Let $\pose$ denotes a vector representing the pose of an articulated shape, and let $\object\,{:}\,\mathbb{R}^d {\to} \{0,1\}$ denotes an occupancy function defining the exterior and interior of an articulated body.
We are interested in modeling the joint distribution of pose and occupancy, which can be decomposed using the chain rule into a conditional occupancy term, and a pose prior term:
\begin{equation}
    p(\pose, \object) = p(\object \given \pose) \: p(\pose)
    \label{eq:joint}
\end{equation}

%
This paper focuses on building an expressive model of $p(\object \given \pose)$, that is, occupancy conditioned on pose.
\Figure{notation} illustrates this problem for $d{=}2$, and clarifies the notation.
There is extensive existing research on pose priors $p(\pose)$ for human bodies and other articulated objects~\cite{htrack,bogo2016keep,kanazawa_cvpr18}.
Our work is orthogonal to such prior models, 
and any parametric or non-parametric $p(\pose)$ 
 can be combined with our $p(\object\given\pose)$ to obtain the joint distribution $p(\pose, \object)$.
We delay the discussion of pose priors until \Section{prior},
where we define a particularly simple prior that nevertheless supports sophisticated tracking of moving humans.

In what follows we describe different ways of building a pose conditioned occupancy function, denoted $\object_\omega(\x \given \pose)$,
which maps a 3D point $\x$ and a pose $\pose$ onto a real valued occupancy value.
Our goal is to learn a parametric occupancy~$\object_\omega(\x \given \pose)$ that mimics a ground truth occupancy $\object(\x \given \pose)$ as closely as possible, based
on the following probabilistic interpretation:
\begin{equation}
    p(\object \given \pose) \propto \prod_{\x\in \R^d} \exp\{ -(\object_\omega(\x \given \pose) - \object(\x \given \pose))^2\}~,
\end{equation}
where we assume a standard normal distribution around the predicted real valued occupancy $\object_\omega(\x \given \pose)$ to 
score an occupancy $\object(\x \given \pose)$.

We are provided with a collection of $T$ ground-truth occupancies~$\{\object(\x \given \pose_t)\}_{t=1}^T$ associated with $T$ poses.
With a slight abuse of notation, we will henceforth use $\x$ to represent both a vector in $\R^d$, and its $\R^{d+1}$ homogeneous representation $[\x;1]$.
In our formulation, each pose parameter $\pose$ represents a set of $B$ \textit{posed} bones/transformations, i.e., $\theta {\equiv} \{\T_b\}_{b=1}^B$.
To help disambiguate the part-whole relationship, we also assume that for each mesh vertex~$v{\in}\V$, the body part associations $\mathbf{\mathbf{w}}(v)$ are available, where $\mathbf{w}(v){\in}[0,1]^B$ with~$\lVert \mathbf{w}(v) \rVert_1 {=}1$.

Given pose parameters $\pose$, we desire to query the corresponding indicator function $\object(\x \given \pose)$ at a point $\x$.
This task is more complicated than might seem, as in the general setting this operation requires the computation of generalized winding numbers to resolve ambiguous configurations caused by self-intersections and non-necessarily watertight geometry~\cite{winding}.
However, when given a database of poses~$\poses{=}\{\pose_t\}_{t=1}^T$ and corresponding \textit{ground truth} indicator $\{\object(\x \given \pose_t)\}_{t=1}^T$, we can formulate our problem as the minimization of the objective:
\begin{equation}
\mathcal{L}_\text{occupancy}(\omega) ~{=}~ \sum_{\pose \in \poses} \expect{\x \sim p(\x)}{ \left(\object(\x \given \pose) - \object_\omega(\x \given \pose) \right)^{2}}
\label{eq:rec_loss}
\end{equation}
where $p(\x)$ is a density representing the sampling distribution of points in $\mathbb{R}^d$~(\Section{techdetails}) and $\object_\omega$ is a neural network with parameters $\omega$ that represents our \textit{neural articulated shape approximator}.
We adopt a sampling distribution $p(\x)$ that randomly samples in the volume surrounding a posed character, along with additional samples in the vicinity of the deformed surface.

\begin{figure*}[t]
\centering
\begin{overpic} 
[width=\stylelinewidth]
{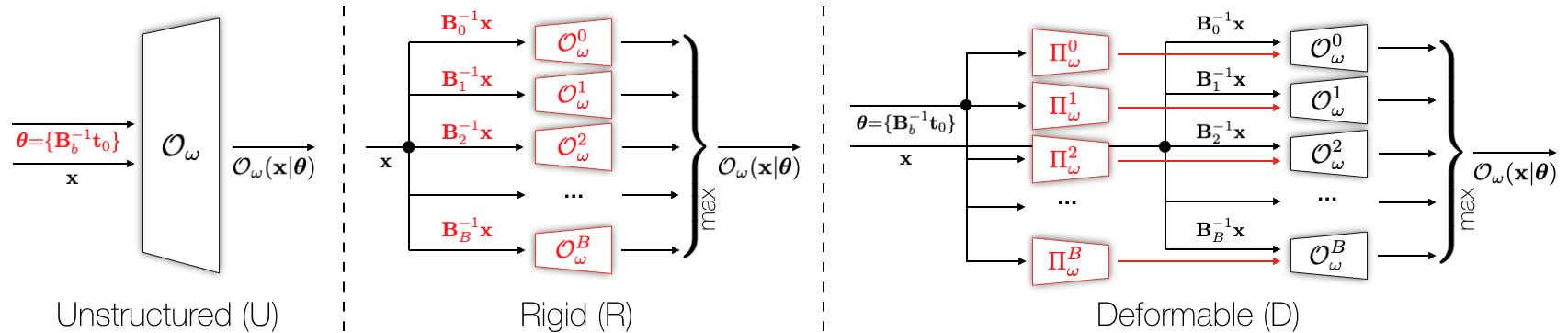}
\end{overpic}
\vspace{-1.5em}
\caption{
The three architectures for $p(\object \given \pose)$.
The unstructured model employs a global MLP conditioned on pose, the rigid model expresses geometry as a composition of $B$ \textit{rigid} elements, while the deformable model via a composition of $B$ \textit{deformable} elements; we highlight the differences between models in red.
}
\label{fig:outline}
\end{figure*}

\section{Pose conditioned occupancy $\object(\x \given \pose)$}
\begin{figure*}[t!]
\centering
\includegraphics[width=\linewidth]{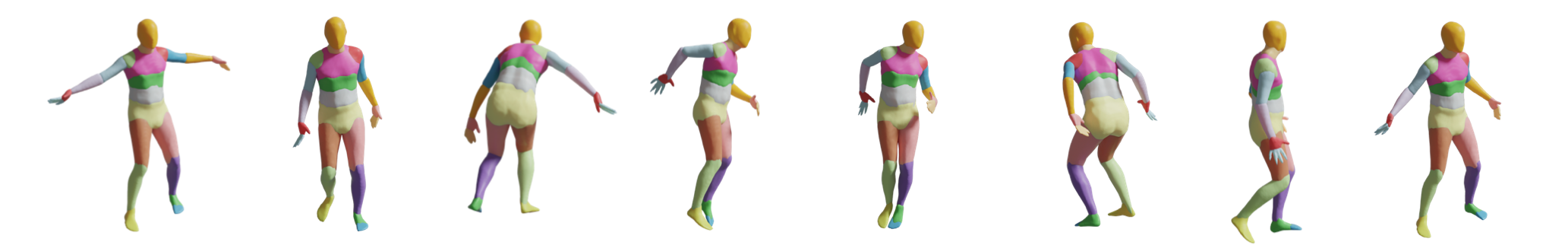}
\vspace{-2em}
\caption{
Our NASA representation models an articulated object as a collection of \textit{deformable} components.
The shape of each component is controlled by the pose of the subject, in a way that take inspiration from pose-space correctives~\cite{posespacedeformation}.
}
\label{fig:decomposition}
\end{figure*}

We investigate several neural architectures for the problem of articulated shape approximation; see~\Figure{outline}.
We start by introducing an unstructured architecture (U) in \Section{unstructured}.
This baseline variant does not explicitly encode the knowledge of articulated deformation.
However, typical articulated deformation models~\cite{smpl} express deformed mesh vertices ${\V}$ reusing the information stored in rest vertices~$\rest{\V}$. 
Hence, we can assume that computing the function $\object(\x | \pose)$ in the deformed pose can be done by reasoning about the information stored at rest pose~$\object(\x | \rest\pose)$.
Taking inspiration from this observation, we investigate two different architecture variants, one that models geometry via a \textit{piecewise-rigid} assumption~(\Section{rigid}),
and one that relaxes this assumption and employs a \textit{quasi-rigid} decomposition, where the shape of each element can deform according to the pose~(\Section{deformable}); see~\Figure{decomposition}.


\subsection{Unstructured model -- ``U''}
\label{sec:unstructured}
Recently, a series of papers~\cite{imnet,deepsdf,occnet} tackled the problem of modeling occupancy across shape datasets as~ $\object_\omega(\x \given \shape)$, where $\shape$ is a latent code learned to encode the shape.
These techniques employ deep and fully connected networks, which one can adapt to our setting by replacing the shape $\shape$ with pose parameters $\pose$, and using a neural network that takes as input $[\x, \pose]$.
%
%
Leaky ReLU activations are used for inner layers of the neural net and a sigmoid activation is used for the final output so that the occupancy prediction lies in the $[0,1]$ range.

To provide pose information to the network, one can simply concatenate the set of affine bone transformations to the query point to obtain $[\point, \{\T_b\} ]$ as the input. This results in an input tensor of size $3{+}16{\times}B$.
Instead, we propose to represent pose as $\{ \T_b^{-1} \mathbf{t}_0 \}$, where $\mathbf{{t}}_0$ is the translation vector of the \textit{root} bone in homogeneous coordinates, resulting in a smaller input of size $3{+}3{\times}B$; we ablate this choice against other alternatives in the \textbf{supplementary material}.
%
Our {unstructured} baseline takes the form:
\begin{equation}
\object_\omega(\x \given \pose) =
\MLP_\omega(\point, \underbrace{\{ \T_b^{-1} \mathbf{t}_0 \}}_\text{pose})
\label{eq:unstructured}
\end{equation}
%


\subsection{Piecewise rigid model -- ``R''}
\label{sec:rigid}
The simplest structured deformation model for articulated objects assumes objects can be represented via a \textit{piecewise rigid} composition of elements;~e.g.~\cite{hadjust,melax}:
\begin{align}
\object(\x \given \pose) = \max_b \{ \object^b(\x \given \pose) \}
\end{align}
We observe that if these elements are related to corresponding rest-pose elements through the rigid transformations~$\{\T_b\}$, then it is possible to \textit{query} the corresponding rest-pose indicator as:
\begin{align}
\object_\omega(\x \given \pose) = \max_b 
\{ \rest\object^{b}_\omega(\T_b^{-1} \point) \} 
\label{eq:piecewise}
\end{align}
where, similar to \eq{unstructured}, we can represent each of components via a \textit{learnable} indicator~$\rest\object^b_\omega(.){=}\MLP^b_\omega(.)$.
This formulation assumes that the local shape of each learned bone component stays \textit{constant} across the range of poses when viewed from the corresponding coordinate frame, which is only a crude approximation of the deformation in realistic characters, and other deformable shapes.

\subsection{Piecewise deformable model -- ``D''}
\label{sec:deformable}
We can generalize our models by combining the model of~\eq{unstructured} to the one in \eq{piecewise}, hence allowing the shape of each element to be \textit{adjusted} according to pose:
\begin{align}
\object_\omega(\x \given \pose) = \max_b 
\{ \rest\object^{b}_\omega( \underbrace{\T_b^{-1} \point}_\text{query} \given \pose) \} 
\end{align}
Similar to $\eq{piecewise}$ we use a \textit{collection} of learnable indicator functions in rest pose~$\{ \object_\omega^b \}$, and to encode pose conditionals we take inspiration from \eq{unstructured}.
More specifically, we express our model as:
\begin{align}
\object_\omega(\x \given \pose) = \max_b 
\{ \rest\object^{b}_\omega(\T_b^{-1} \point, \underbrace{\Pi_\omega^b\left[\{\T_b^{-1}{\mathbf{t}}_0\}\right]}_\text{part-specific pose}) \} 
\label{eq:deformable}
\end{align}
Similarly to \eq{piecewise}, we model $\rest\object^b_\omega(.)$ via dense layers $\MLP^b_\omega: \R^{3+D} {\rightarrow} \R$.
The operator $\Pi_\omega^b : \R^{B{\times}3} {\rightarrow} \R^D$ is a learnable \textit{linear} subspace projection -- one per each bone $b$.
This choice is driven by the intuition that in typical skinned deformation models only \textit{small subset} of the coordinate frames affect the deformation of a part.
We employ $D{=}4$ throughout, see ablations in the supplementary material.
Our experiments reveal that this bottleneck greatly improves generalization.


\subsection{Technical details}
\label{sec:techdetails}
The overall training loss for our model is: 
\typeout{ CHANGED ALIGN* -> ALIGN HERE }
\begin{align}
\mathcal{L}(\omega) =
\mathcal{L}_\text{occupancy}(\omega)
+ \lambda \mathcal{L}_\text{weights}(\omega)
\label{eq:train}
\end{align}
where $\lambda{=}5e^{-1}$ was found through hyper-parameter tuning.
We now detail the weights auxiliary loss, the architecture backbones, and the training procedure.


\paragraph{Auxiliary loss -- skinning weights}
As most deformable models are equipped with skinning weights, we exploit this additional source of information to facilitate learning of the part-based models~(i.e.~``R'' and~``D'').
We label each mesh vertex $\vertex$ with the index of the corresponding highest skinning weight value $b^*(v){=}\argmax_b w(v)[b]$, and use the loss:
\begin{equation}
\mathcal{L}_\text{weights}(\omega) = 
\tfrac{1}{V}
\tfrac{1}{B}
\sum_{\pose \in \poses} 
\sum_{\vertex}
\sum_{b}
\left(
\rest\object^b_\omega(\vertex \given \pose) - \mathcal{I}_b(\vertex)
\right)^{\!\!2}
\label{eq:skinningw-loss}
\end{equation}
where $\mathcal{I}_{b}(\vertex){=}0.5$ when $b{=}b^*$, and $\mathcal{I}_{b}(\vertex){=}0$ otherwise -- recall that by convention the $0.5$~level set is the surface represented by the occupancy function. 
Without such a loss, we could end up in the situation where a single (deformable) part could end up being used to describe the entire deformable model, and the trivial solution~(zero) would be returned for all other parts.



\paragraph{Network architectures}
To keep our experiments comparable across baselines, we use the same network architecture for all the models while varying the \textit{width} of the layers.
The network backbone is similar to DeepSDF~\cite{deepsdf}, but simplified to 4 layers.
Each layer has a residual connection, and uses the Leaky ReLU activation function with the leaky factor 0.1.
All layers have the \textit{same} number of neurons, which we set to $960$ for the unstructured model and $40$ for the structured ones.
For the piecewise~\eq{piecewise} and deformable~\eq{deformable} models the neurons are distributed across $B{=}24$ different channels (note $B {\times} 40=960$).
Similar to the use of grouped filters/convolutions~\cite{alexnet,groups}, such a structure allows for significant performance boosts compared to unstructured models~\eq{unstructured}, as the different branches can be executed in \textit{parallel} on separate compute devices.


\paragraph{Training}
All models are trained with the Adam optimizer, with batch size~$12$ and learning rate $1e-4$.
For better gradient propagation, we use \textit{softmax} whenever a max was employed in our expressions.
For each optimization step, we use $1024$ points sampled uniformly within the bounding box and $1024$ points sampled near the ground truth surface. We also sample $2048$ vertices out of $6890$ mesh vertices at each step for $\mathcal{L}_{weights}$.
The models are trained for $200$K iterations for approximately $6$ hours on a single NVIDIA Tesla V100.

\section{Dense articulated tracking}
\label{sec:densetracking}
Following the probabilistic interpretation of \Section{nasa}, we introduce an application of NASA to dense articulated 3D \textit{tracking}; see~\cite{regcourse_sgp18}.
Note that this section does not claim to beat the state-of-the-art in tracking of deformable objects~\cite{shen2020phong,taylor2017articulated,tkach2017online}, but rather it seeks to show \textit{how} neural occupancy functions can be used effectively in the development of dense tracking techniques.
Taking the negative log of the joint probability in \eq{joint}, the tracking problems can be expressed as the minimization
of a pair of energies~\cite{regcourse_sgp18}:
\begin{equation}
\argmin_{\pose^{(t)}} \:\: E_\text{fit}( \depth^{(t)}, \pose^{(t)} ) + E_\text{prior}(\pose^{(t)})
\end{equation}
where $\depth{=}\{\point_n\}_{n=1}^N$ is a point cloud in $\R^d$, and the superscript $(t)$ indicates the point cloud and the pose associated with the $t^\text{th}$ frame.
The optimization for~$\pose^{(t)}$ is initialized with the minimizer computed at frame~$(t{-}1)$.
We also assume~$\pose^{(0)}$ is provided as ground truth, but discriminative models could also be employed to obtain an initialization~\cite{kinect,kanazawa_cvpr18}.
In what follows, we often drop the ${(t)}$ superscript for clarity of notation.
We now discuss different aspects of this problem when an implicit representation of the model is used, including the implementation of fitting~(\Section{fitting}) and prior~(\Section{prior}) energies, as well as details about the iterative optimization scheme~(\Section{iterative}).


\subsection{Fitting energy}
\label{sec:fitting}
If we could compute the Signed Distance Function~(SDF) $\SDF$ of an occupancy $\object$ at a query point $\point$,
then the fitness of $\object$ to input data could be measured as:
\begin{equation}
E_\text{fit}(\depth, \pose) = \sum_{\point \in \depth} \| \SDF(\point \given \object, \pose) \|^2
\label{eq:sdftrack}
\end{equation}
The time complexity of computing SDF from an occupancy $\object$ that is discretized on a grid
is \textit{linear} in the number of voxels~\cite{felzenszwalb}.
However, the number voxels grows as $O(n^d)$, making naive SDF computation impractical for high resolutions~(large $n$) or high dimensions~(in practice, $d \text{\scriptsize $\geq$} 3$ is already problematic).
Spatial acceleration data structures~(kdtrees and octrees) are commonly employed, but these data structures still require an overall $O(n \log(n))$ pre-processing~(where $n$ is the number of polygons), and they need to be re-built at every frame (as $\pose$ changes), and do not support implicit representations.

Recently, Dou et al.~\cite{fusion4d} proposed to smooth an occupancy function with a Gaussian blur kernel to \textit{approximate} 
$\SDF$ in the near field of the surface. 
Following this idea, our fitting energy can be re-expressed as:
\begin{align}
E_\text{fit}(\depth, \pose) = \sum_{\point \in \depth} \| \normal_{0,\sigma^2} \circledast \object(\point \given \pose) - 0.5 \|^2
\label{eq:occtrack}
\end{align}
where $\normal_{0,\sigma^2}$ is a Gaussian kernel with a zero mean and a variance $\sigma^2$, and $\circledast$ is the convolution operator.
This approximation is suitable for tracking, as large values of distance should be associated with \textit{outliers} in a registration optimization~\cite{sparseicp}, and therefore ignored.
Further, this approximation can be explained via the algebraic relationship between heat kernels and
distance functions~\cite{crane2013geodesics}.
Note that we \textit{intentionally} use $\object$ instead of $\object_\omega$, as what follows is applicable to \textit{any} implicit representation, not just our neural occupancy $\object_\omega$.

Dou~et~al.~\cite{fusion4d} used \eq{occtrack} being given a voxelized representation of $\object$, and relying on GPU implementations to efficiently compute 3D convolutions $\circledast$.
To circumvent these issues, we re-express the convolution via stochastic sampling:
\newcommand{\gaussian}{g}
\newcommand{\otherpoint}{\mathbf{s}} 
\begin{align}
\object(\point \given \pose) \circledast \normal_{0,\sigma^2}
&= \int \object(\otherpoint \given \pose) \gaussian(\point - \mathbf{s} | 0, \sigma^2) \: d \mathbf{s} &\text{(definition of convolution)}\\
&= \int \object(\otherpoint \given \pose) \gaussian(\mathbf{s}-\point| 0, \sigma^2) \: d \mathbf{s} &\text{(symmetry of Gaussian)} \\
&= \int \object(\otherpoint \given \pose) \gaussian(\mathbf{s} | \point, \sigma^2) \: d \mathbf{s} &\text{(definition of Gaussian)}\\
&= \expect{\mathbf{s} \sim \normal_{\point, \sigma^2} }{\object(\mathbf{s} \given \pose)} &\text{(definition of expectation)}
\label{eq:convexp}
\end{align}
Overall, \Equation{convexp} allows us to design a tracking solution that directly operates on occupancy functions, \textit{without} the need to compute signed distance functions~\cite{htrack}, closest points~\cite{regcourse_sgp18}, or 3D convolutions~\cite{fusion4d}.
It further provides a direct cost/accuracy control in terms of the number of samples used to approximate the expectation in~\eq{convexp}.
However, the gradients $\nabla_{\pose}$ of \eq{convexp} also need to be available -- we achieve this by applying the \textit{re-parameterization} trick~\cite{reparam} to~\eq{convexp}:
\begin{align}
\nabla_{\pose}
\left[ \expect{\mathbf{s} \sim \normal_{\point, \sigma^2} }{\object(\mathbf{s} \given \pose)} \right] = 
\expect{\mathbf{s} \sim \normal_{0,1}}{\nabla_{\pose} \object(\point + \sigma \mathbf{s} \given \pose)}
\label{eq:random}
\end{align}


\subsection{Pose prior energy}
\label{sec:prior}
An issue of generative tracking is that once the model is too far from the target (e.g. fast motion) there will be no proper gradient to correct it.
If we directly optimize for transformation without any constraints, there is a high chance that the model will degenerate into such a case.
To address this, we impose a prior:
\begin{align}
    E_\text{prior}(\pose{=}\{\T_b\}) =\!\!\!\! \sum_{(b_1, b_2) \in \mathcal{E}} \left\| (\rest\trans_{b_2} - \rest\trans_{b_1}) - \T_{b_1}^{-1} \trans_{b_2} \right\|_2^2
\end{align}
where $\mathcal{E}$ is the set of directed edges $(b_1, b_2)$ on the pre-defined directed rig with $b_1$ as the parent, and recall $\mathbf{t}_b$ is the translation vector of matrix $\T_b$.
One can view this loss as aligning the vector pointing to $t_{b_2}$ at run-time with the vector at rest pose, i.e. $(\rest\trans_{b_2} - \rest\trans_{b_1})$.
We emphasize that more sophisticated priors exist, and could be applied, including employing a hierarchical skeleton~\cite{htrack}, or modeling the density of joint angles~\cite{bogo2016keep}.
The simple prior used here is chosen to highlight the effectiveness of our neural occupancy model \textit{independent} of such priors.


\subsection{Iterative optimization}
\label{sec:iterative}
One would be tempted to use the gradients of \eq{occtrack} to track a point cloud via \textit{iterative} optimization.
However, it is known that when optimizing rotations \textit{centering} the optimization about the current state is heavily advisable~\cite{regcourse_sgp18}.
Indexing time by $(t)$ and given the update rule $\pose^{(t)} {=} \pose^{(t-1)} {+}\, \Delta\pose^{(t)}$, the iterative optimization of \eq{occtrack} can be expressed as:
\begin{align}
\argmin_{\Delta\pose^{(t)}} \sum_{\point \in \depth^{(t)}}
\left\| \expect{\mathbf{s} \sim \normal_{\point, \sigma^2} }{ \object_\omega(\mathbf{s} \given \pose^{(t-1)} + \Delta\pose^{(t)})} - 0.5 \right\|^2
\label{eq:iteropt}
\end{align}
where in what follows we omit the index $(t)$ for brevity of notation.
As the pose~$\pose$ is represented by matrices, we represent the transformation differential as:
\begin{align}
\object_\omega(\x \given \pose+\Delta\pose) &= \object_\omega(\x \given \{(\T_b \Delta \T_b)^{-1}\}) = 
\object_\omega(\x \given \{ \Delta \T_b^{-1} \T_b^{-1} \}),
\label{eq:post}
\end{align}
%
resulting in the optimization:
\begin{align}
\argmin_{\{\Delta \T_b^{-1}\}} \:\: \sum_{\point \in \depth} \left\| 
\expect{\mathbf{s} \sim \normal_{\point, \sigma^2} }{\object_\omega(\mathbf{s} \given \{ \Delta \T_b^{-1} \T_b^{-1} \})} - 0.5 \right\|^2
\label{eq:tracking}
\end{align}
where we parameterize the rotational portion of elements in the collection $\{\Delta \T_b^{-1}\}$ by two (initially orthogonal) vectors~\cite{rotations}, and re-orthogonalize them \textit{before} inversion within each optimization update $\T_b^{(i+1)} {=} \T_b^{(i)} (\Delta \mathbf{C}_b^{(i)})^{-1}$, where $\Delta\mathbf{C}_b{=}\Delta\T_b^{-1}$ are the quantities the solver optimizes for.
In other words, we optimize for the \textit{inverse} of the coordinate frames in order to avoid back-propagation through matrix inversion.
\section{Results and discussion}
\label{sec:evaluation}
We describe the training data~(\Section{data}), quantitatively evaluate the performance of our neural 3D representation on several datasets~(\Section{reconstruction}), as well as demonstrate its usability for tracking applications~(\Section{trackingresults}).
We conclude by contrasting our technique to recent methods for implicit-learning of geometry~(\Section{relatedplus}).
Ablation studies validating \textit{each} of our technical choices can be found in the
\textbf{supplementary material}.

\subsection{Training data}
\label{sec:data}
Our training data consists of 
sampled indicator function values,  transformation frames (``bones'') per pose, and skinning weights.
The samples used for training~\eq{rec_loss} come from two sources (each comprising a total of $100,000$ samples):
\CIRCLE{1}~we randomly sample points uniformly within a bounding box scaled to 110\% of its original diagonal dimension;
\CIRCLE{2}~we perform Poisson disk sampling on the surface, and randomly displace these points with isotropic normal noise with $\sigma{=}.03$.
The ground truth indicator function at these samples are computed by casting randomized rays and checking the \textit{parity} (i.e. counting the number of intersections) -- generalized winding numbers~\cite{winding} or sign-agnostic losses~\cite{sal} could also be used for this purpose.
The test reconstruction performance is evaluated by comparing the predicted indicator values against the ground truth samples on the full set of $100,000$ samples.
We evaluate using mean Intersection over Union (IoU), Chamfer-L1~\cite{chamfer} and F-score (F$\%$)~\cite{fscore} with a threshold set to $0.0001$.
The meshes are obtained from the ``DFaust''~\cite{dfaust:CVPR:2017} and ``Transitions'' sub-datasets of AMASS~\cite{amass}, as detailed in~\Section{reconstruction}.


\begin{figure*}[t]
\centering
\begin{minipage}{.4\columnwidth}
\centering
\resizebox{\linewidth}{!}{
\begin{tabular}{c|ccc}
\toprule
Model & mIoU$\uparrow$ & Chamfer~L1$\downarrow$ & F\%$\uparrow$ \\ 
\midrule
{\color{red} \textbf{U}} & .702 & .00631 & 46.15 \\
{\color{purple} \textbf{R}} & .932 & .00032 & 93.94 \\
{\color{corn} \textbf{D}} & \textbf{.959} & \textbf{.00004} & \textbf{98.54} \\
\bottomrule
\end{tabular}
} 
\vspace{-1em}
\captionof{table}{AMASS / DFaust}
\label{tbl:dfaust}
\end{minipage}
\hspace{.09\columnwidth}
\begin{minipage}{.4\columnwidth}
\centering
\resizebox{\linewidth}{!}{
\begin{tabular}{c|ccc}
\toprule
Model & mIoU$\uparrow$ & Chamfer~L1$\downarrow$ & F\%$\uparrow$ \\ 
\midrule
{\color{red} \textbf{U}} & .520 & .01057 & 26.83 \\
{\color{purple} \textbf{R}} & .936 & .00006 & 96.71 \\
{\color{corn} \textbf{D}} & \textbf{.965} & \textbf{.00002} & \textbf{99.42} \\
\bottomrule
\end{tabular}
} 
\vspace{-1em}
\captionof{table}{AMASS / Transitions}
\label{tbl:transitions}
\end{minipage}
\\[.5em]
\begin{minipage}{\columnwidth}
\centering
\begin{overpic} 
[width=\columnwidth]
{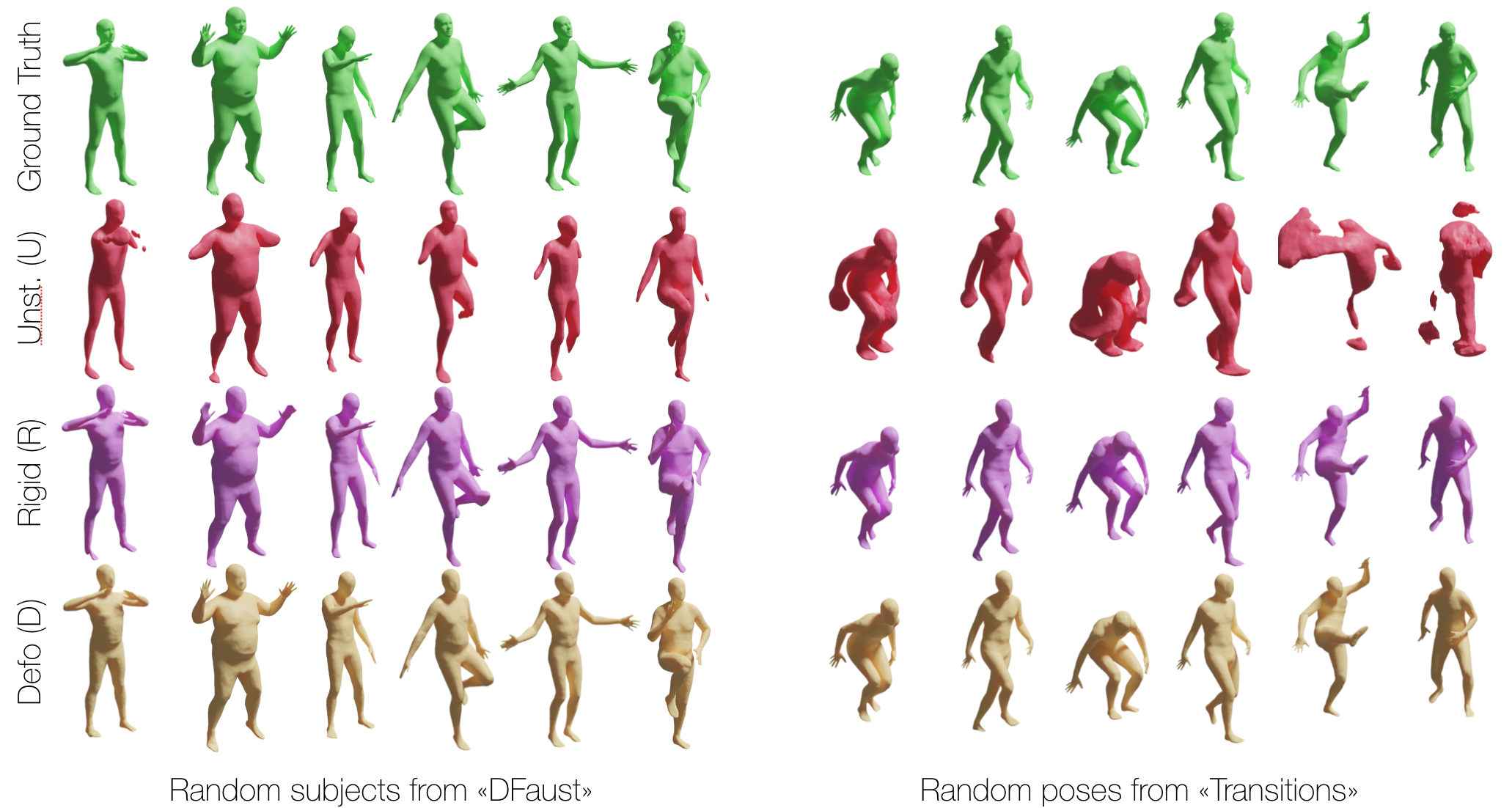}
\end{overpic}
\vspace{-2em}
\caption{The qualitative performance of our three models in reconstructing the occupancy function on the (left) DFaust and (right) Transitions dataset.}
\label{fig:reconstruction}
\end{minipage}
\end{figure*}
\subsection{Reconstruction}
\label{sec:reconstruction}
%
We employ the ``DFaust'' portion of the AMASS dataset to verify that our model can be used effectively \textit{across} different subjects.
This dataset contains $10$ subjects, $10$ sequences/subject, and ${\approx}300$ frames/sequence on average.
We train $100$ different models by optimizing~\eq{train}: for each subject we use $9$ sequences for training, leaving one out for testing to compute our metrics. 
We average these metrics across the $100$ runs, and report these in~\Table{dfaust}.
Note how learning a deformable model via decomposition provides \textit{striking} advantages, as quantified by the fact that the rigid~(R) baseline is consistently better than the unstructured~(U) baseline under \textit{any} metric -- a $\mathbf{+49\%}$ in F-score.
Similar improvements can be noticed by comparing the rigid (R) to the deformable (D) model, where the latter achieves an additional $\mathbf{+5\%}$ in F-score.
\Figure{reconstruction} (second row) gives a qualitative visualization of how the unstructured models struggles in generalizing to poses that are sufficiently different from the ones in the training set.

We employ the ``Transitions'' portion of the AMASS dataset to further study the performance of the model when more training data and a larger diversity of motions) is available for a \textit{single} subject.
This dataset contains $110$ sequences of one individual, with ${\approx}1000+$ frames/sequence.
We randomly sample ${\approx}250$ frames from each sequence, randomly select $80$ sequences for training, and keep the remaining $30$ sequences for testing; see our \textbf{supplementary material}.
Results are shown in~\Table{transitions}.
The conclusions are analogous to the ones we made from DFaust.
Further, note that in this more difficult dataset containing a larger variety of more complex motions, the unstructured model struggles even more significantly ($U{\rightarrow}R$: $\mathbf{+70\%}$ in F-score).
As the model is exposed to more poses compared to DFaust, the reconstruction performance is also improved.
Moving from DFaust to Transitions results in a  $\mathbf{+1\%}$ in F-score for the deformable model.


\begin{figure*}[t]
\centering
\begin{minipage}{\columnwidth}
\centering
\begin{overpic} 
[width=\columnwidth]
{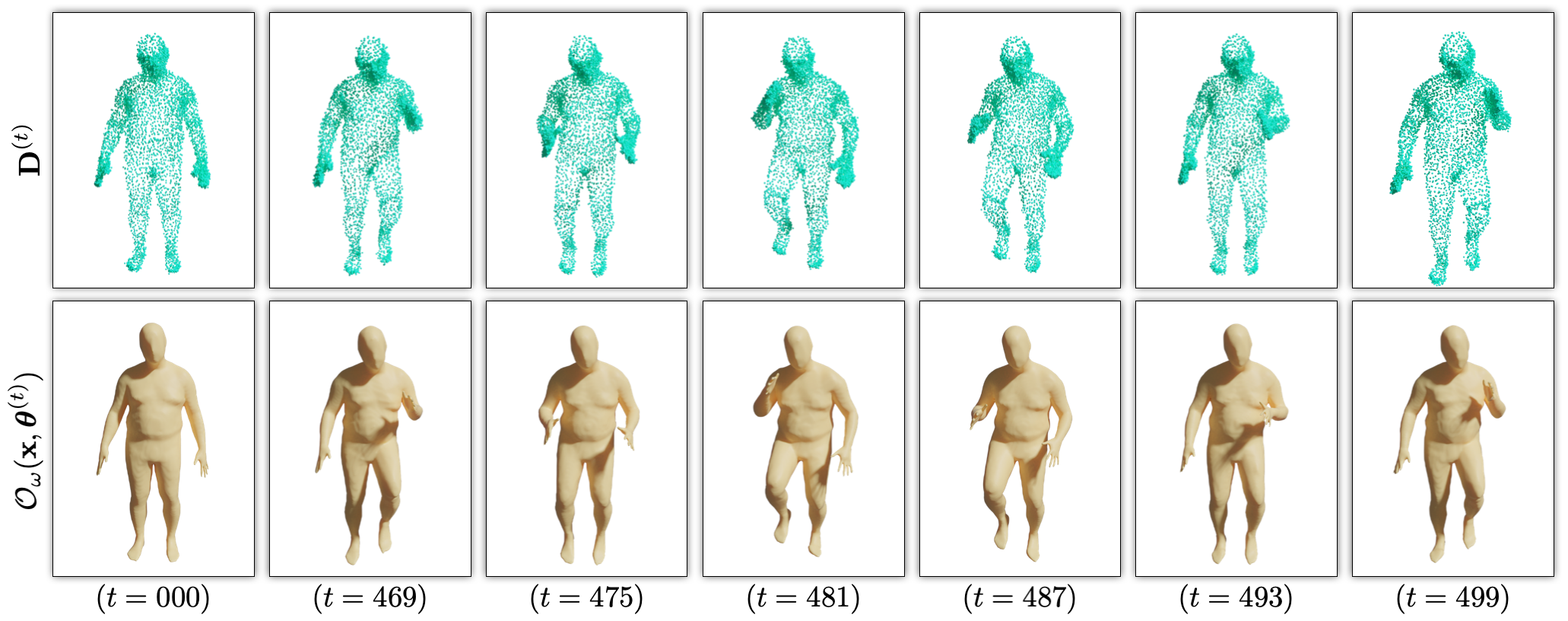}
\end{overpic}
\vspace{-2em}
\caption{
A few frames of our neural model \textit{tracking} the point cloud of the DFaust ``hard'' (02-09) sequence; these results can be better appreciated in our \textbf{video}.
}
\label{fig:tracking}
\end{minipage} 
\\[1em]
\begin{minipage}{.49\columnwidth}
\centering
\resizebox{\linewidth}{!}{
\begin{tabular}{ccc|ccc}
\toprule
$p(\mathcal{O} \given \pose)$ & $p(\pose)$ & $\circledast$ & mIoU$\uparrow$ & Chamfer~L1$\downarrow$ & F\%$\uparrow$ \\ 
\midrule
U & \cmark & \cmark & .845 & .00383 & 61.63 \\
D & \cmark & \cmark & \textbf{.968} & \textbf{.00004} & \textbf{99.08} \\
\midrule
oracle & -- & -- & .976 & .00004 & 99.01\\
\bottomrule
\end{tabular}
} 
\vspace{-1em}
\captionof{table}{DFaust ``easy'' (00-01)}
\label{tbl:dfaust_easy}
\end{minipage}
\begin{minipage}{.49\columnwidth}
\centering
\resizebox{\linewidth}{!}{
\begin{tabular}{ccc|ccc}
\toprule
$p(\mathcal{O} \given \pose)$ & $p(\pose)$ & $\circledast$ & mIoU$\uparrow$ & Chamfer~L1$\downarrow$ & F\%$\uparrow$ \\ 
\midrule
U & \cmark & \cmark & .686 & .00700 & 50.63 \\
D & \cmark & \cmark & \textbf{.948} & \textbf{.00006} & \textbf{96.48} \\
\midrule
oracle & -- & -- & .959 & .00006 & 96.80\\
\bottomrule
\end{tabular}
} 
\vspace{-1em}
\captionof{table}{DFaust ``hard'' (02-09)}
\label{tbl:dfaust_hard}
\end{minipage}
\end{figure*}
\subsection{Tracking}
\label{sec:trackingresults}
We validate our tracking technique on two sequences from the DFaust dataset;~see \Figure{tracking}.
Note these are test sequences, and were \textit{not} used to train our model.
The prior $p(\pose)$ (\Section{prior}) and the stochastic optimization $\circledast$ (\Section{fitting}) can be applied to both unstructured~(U) and structured~(D) representations, with the latter leading to significantly better tracking performance.
The quantitative results reported for the ``easy''~(\Table{dfaust_easy}) and ``hard''~(\Table{dfaust_hard}) tracking sequences are best understood by watching our \textbf{supplementary video}.
It is essential to re-state that we are not trying to beat traditional baselines, but rather seek to \textit{illustrate} how NASA, once trained, can be readily used as a 3D representation for classical vision tasks.
For the purpose of this illustration, we use only noisy panoptic point clouds (i.e. complete~\cite{fusion4d} rather than incomplete~\cite{htrack} data), 
and do not use any discriminative per-frame re-initializer as would typically be employed in a contemporary tracking system.

\subsection{Discussion}
\label{sec:relatedplus}
\vspace{-.05in}
The recent success of neural implicit representations of geometry, introduced by~\cite{imnet,deepsdf,occnet}, has heavily relied on the fact that the geometry in ShapeNet datasets~\cite{shapenet} is \textit{canonicalized}: scaled to unit ranges and consistently oriented. 
Research has highlighted the importance of expressing information in a canonical frame~\cite{nocs},
and one could interpret our method as a way to achieve this within the realm of articulated motion.
To understand the shortcomings of unstructured models, one should remember that as an object moves,
much of the local geometric details remain \textit{invariant} to articulation
(e.g. the geometry of a wristwatch does not change as you move your arm).
However, unstructured pose conditioned models are forced to \textit{memorize} these details in any pose they seek to reconstruct.
Hence, as one evaluates unstructured models \textit{outside} of their training manifold, their performance \textit{collapses} -- as quantified by the $+49\%$ performance change as we move from unstructured to rigid models; see~\Table{dfaust}.
One could also argue that given sufficient capacity, a neural network \textit{should} be able to learn the
\textit{concept} of coordinate frames and transformations.
However, multiplicative relationships between inputs (e.g.~dot products) are difficult to learn for neural networks~\cite[Sec.~2.3]{moment}.
As changes of coordinate frames are nothing but collections of dot products, one could use this reasoning to justify the limited performance of
unstructured models.
We conclude by clearly contrasting our method, targeting the modeling of $\object( \x \given \pose)$ to those that address shape completion $\object(\x \given \depth)$~\cite{dsif,disn,pifu,sal}.
In contrast to these, our solution, to the best of our knowledge, represents the first attempt to create a ``neural implicit rig'' -- from a computer graphics perspective -- for articulated deformation modeling.

One limitation of our work is the reliance on $\{\tilde\T_b\}$, which could be difficult to obtain in in-the-wild settings, as well as skinning weights to guide the part-decomposition; how to automatically regress these quantities from raw observations is an open problem.
Our model is also currently limited to \textit{individual} subjects, and to be competitive to mesh-based models one would also have to learn identity parameters~(i.e.~the $\shape$ parameters of SMPL~\cite{smpl}).
Finally, our representation currently fails to capture high frequency features (e.g.~see the geometric details of the face region in~\Figure{reconstruction}); however, recent research on implicit representations~\cite{tancik2020fourierfeat,sitzmann2020implicit} can likely mitigate this issue.
\section{Conclusions}
\label{sec:conclusions}
\vspace{-.05in}
We introduce a novel neural representation of a particularly important class of 3D objects: \textit{articulated} bodies.
We use a structured neural occupancy approach, enabling both direct occupancy queries and deformable surface representations that are competitive with classic hand-crafted mesh representations.
The representation is fully differentiable, and enables tracking of realistic articulated bodies -- traditionally a complex task -- to be almost \textit{trivially} implemented.
Crucially, our work demonstrates the value of incorporating a task-appropriate inductive bias into the neural architecture. 
By acknowledging and encoding the quasi-rigid part structure of articulated bodies, we represent this class of objects with higher quality, and significantly better generalization.

\clearpage
\bibliographystyle{splncs04}
\bibliography{macros,main}

\begin{thebibliography}{10}
\providecommand{\url}[1]{\texttt{#1}}
\providecommand{\urlprefix}{URL }
\providecommand{\doi}[1]{https://doi.org/#1}

\bibitem{anguelovArticulated}
Anguelov, D., Koller, D., Pang, H.C., Srinivasan, P., Thrun, S.: Recovering
  articulated object models from 3d range data. In: Uncertainty in Artificial
  Intelligence (2004)

\bibitem{sal}
Atzmon, M., Lipman, Y.: Sal: Sign agnostic learning of shapes from raw data.
  arXiv preprint arXiv:1911.10414  (2019)

\bibitem{sig18skinning}
Bailey, S.W., Otte, D., Dilorenzo, P., O'Brien, J.F.: Fast and deep deformation
  approximations. SIGGRAPH  (2018)

\bibitem{bogo2016keep}
Bogo, F., Kanazawa, A., Lassner, C., Gehler, P., Romero, J., Black, M.J.: Keep
  it {SMPL}: Automatic estimation of 3d human pose and shape from a single
  image. In: ECCV (2016)

\bibitem{dfaust:CVPR:2017}
Bogo, F., Romero, J., Pons-Moll, G., Black, M.J.: Dynamic {FAUST}: Registering
  human bodies in motion. In: {IEEE} Conf. on Computer Vision and Pattern
  Recognition (CVPR) (2017)

\bibitem{sparseicp}
Bouaziz, S., Tagliasacchi, A., Pauly, M.: Sparse iterative closest point. In:
  SGP (2013)

\bibitem{shapenet}
Chang, A.X., Funkhouser, T., Guibas, L., Hanrahan, P., Huang, Q., Li, Z.,
  Savarese, S., Savva, M., Song, S., Su, H., et~al.: Shapenet: An
  information-rich 3d model repository. arXiv:1512.03012  (2015)

\bibitem{baenet}
Chen, Z., Yin, K., Fisher, M., Chaudhuri, S., Zhang, H.: Bae-net: Branched
  autoencoder for shape co-segmentation. In: ICCV (2019)

\bibitem{imnet}
Chen, Z., Zhang, H.: Learning implicit fields for generative shape modeling.
  CVPR  (2019)

\bibitem{chibane20ifnet}
Chibane, J., Alldieck, T., Pons-Moll, G.: Implicit functions in feature space
  for 3d shape reconstruction and completion. In: CVPR (2020)

\bibitem{crane2013geodesics}
Crane, K., Weischedel, C., Wardetzky, M.: Geodesics in heat: A new approach to
  computing distance based on heat flow. ACM TOG  (2013)

\bibitem{cvxnet}
Deng, B., Genova, K., Yazdani, S., Bouaziz, S., Hinton, G., Tagliasacchi, A.:
  Cvxnet: Learnable convex decomposition. CVPR  (2020)

\bibitem{cerberus}
Deng, B., Kornblith, S., Hinton, G.: Cerberus: A multi-headed derenderer.
  arXiv:1905.11940  (2019)

\bibitem{fusion4d}
Dou, M., Khamis, S., Degtyarev, Y., Davidson, P., Fanello, S.R., Kowdle, A.,
  Escolano, S.O., Rhemann, C., Kim, D., Taylor, J., et~al.: Fusion4d: Real-time
  performance capture of challenging scenes. ACM TOG  (2016)

\bibitem{chamfer}
Fan, H., Su, H., Guibas, L.J.: A point set generation network for 3d object
  reconstruction from a single image. In: CVPR (2017)

\bibitem{felzenszwalb}
Felzenszwalb, P.F., Huttenlocher, D.P.: Distance transforms of sampled
  functions. Theory of computing  (2012)

\bibitem{sdmnet}
Gao, L., Yang, J., Wu, T., Yuan, Y.J., Fu, H., Lai, Y.K., Zhang, H.: {SDM-NET}:
  deep generative network for structured deformable mesh. ACM TOG  (2019)

\bibitem{dsif}
Genova, K., Cole, F., Sud, A., Sarna, A., Funkhouser, T.: Deep structured
  implicit functions. CVPR  (2019)

\bibitem{hierarchicalsegment}
de~Goes, F., Goldenstein, S., Velho, L.: A hierarchical segmentation of
  articulated bodies. In: SGP (2008)

\bibitem{atlasnet}
Groueix, T., Fisher, M., Kim, V.G., Russell, B.C., Aubry, M.: Atlasnet: A
  papier-mâché approach to learning 3d surface generation. arXiv preprint
  arXiv:1802.05384  (2018)

\bibitem{jointshapeseg}
Huang, Q., Koltun, V., Guibas, L.: Joint shape segmentation with linear
  programming. ACM TOG  (2011)

\bibitem{groups}
Ioannou, Y., Robertson, D., Cipolla, R., Criminisi, A.: {Deep Roots: Improving
  CNN efficiency with hierarchical filter groups}. In: CVPR (2017)

\bibitem{skinningcourse}
Jacobson, A., Deng, Z., Kavan, L., Lewis, J.: Skinning: Real-time shape
  deformation. In: ACM SIGGRAPH Courses (2014)

\bibitem{winding}
Jacobson, A., Kavan, L., Sorkine-Hornung, O.: Robust inside-outside
  segmentation using generalized winding numbers. ACM TOG  (2013)

\bibitem{skinningmeshanim}
James, D.L., Twigg, C.D.: Skinning mesh animations. SIGGRAPH  (2005)

\bibitem{moment}
Joseph-Rivlin, M., Zvirin, A., Kimmel, R.: Momen(e)t: Flavor the moments in
  learning to classify shapes. In: CVPR Workshops (2019)

\bibitem{kanazawa_cvpr18}
Kanazawa, A., Black, M.J., Jacobs, D.W., Malik, J.: End-to-end recovery of
  human shape and pose. In: CVPR (2018)

\bibitem{reparam}
Kingma, D.P., Welling, M.: Auto-encoding variational bayes. arXiv preprint
  arXiv:1312.6114  (2013)

\bibitem{alexnet}
Krizhevsky, A., Sutskever, I., Hinton, G.: Imagenet classification with deep
  convolutional neural networks. In: NIPS (2012)

\bibitem{skinningdecomp}
Le, B.H., Deng, Z.: Smooth skinning decomposition with rigid bones. ACM TOG
  (2012)

\bibitem{posespacedeformation}
Lewis, J.P., Cordner, M., Fong, N.: Pose space deformation: A unified approach
  to shape interpolation and skeleton-driven deformation. In: SIGGRAPH (2000)

\bibitem{Lin96collisiondetection}
Lin, M.C., Manocha, U.D., Cohen, J.: Collision detection: Algorithms and
  applications (1996)

\bibitem{smpl}
Loper, M., Mahmood, N., Romero, J., Pons-Moll, G., Black, M.J.: {SMPL}: A
  skinned multi-person linear model. SIGGRAPH Asia  (2015)

\bibitem{lorenz2019unsupervised}
Lorenz, D., Bereska, L., Milbich, T., Ommer, B.: Unsupervised part-based
  disentangling of object shape and appearance. arXiv:1903.06946  (2019)

\bibitem{amass}
Mahmood, N., Ghorbani, N., Troje, N.F., Pons-Moll, G., Black, M.J.: {AMASS}:
  Archive of motion capture as surface shapes. ICCV  (2019)

\bibitem{melax}
Melax, S., Keselman, L., Orsten, S.: Dynamics based 3d skeletal hand tracking.
  In: Graphics Interface (2013)

\bibitem{occnet}
Mescheder, L., Oechsle, M., Niemeyer, M., Nowozin, S., Geiger, A.: Occupancy
  networks: Learning 3d reconstruction in function space. arXiv:1812.03828
  (2018)

\bibitem{omran2018NBF}
Omran, M., Lassner, C., Pons-Moll, G., Gehler, P., Schiele, B.: Neural body
  fitting: Unifying deep learning and model based human pose and shape
  estimation. In: International Conference on 3D Vision (3DV) (sep 2018)

\bibitem{deepsdf}
Park, J.J., Florence, P., Straub, J., Newcombe, R., Lovegrove, S.: Deep{SDF}:
  Learning continuous signed distance functions for shape representation. CVPR
  (2019)

\bibitem{pavlakos2018humanshape}
Pavlakos, G., Zhu, L., Zhou, X., Daniilidis, K.: Learning to estimate 3{D}
  human pose and shape from a single color image. In: CVPR (2018)

\bibitem{hadjust}
Remelli, E., Tkach, A., Tagliasacchi, A., Pauly, M.: Low-dimensionality
  calibration through local anisotropic scaling for robust hand model
  personalization. In: ICCV (2017)

\bibitem{pifu}
Saito, S., Huang, Z., Natsume, R., Morishima, S., Kanazawa, A., Li, H.: {PIFu}:
  Pixel-aligned implicit function for high-resolution clothed human
  digitization. In: CVPR (2019)

\bibitem{sametSDSbook}
Samet, H.: Applications of Spatial Data Structures: Computer Graphics, Image
  Processing, and GIS. Addison-Wesley Longman Publishing Co., Inc. (1990)

\bibitem{schmidt2014dart}
Schmidt, T., Newcombe, R.A., Fox, D.: {DART}: Dense articulated real-time
  tracking with consumer depth cameras. Autonomous Robots  \textbf{39}(3)
  (2015)

\bibitem{shen2020phong}
Shen, J., Cashman, T.J., Ye, Q., Hutton, T., Sharp, T., Bogo, F., Fitzgibbon,
  A.W., Shotton, J.: The phong surface: Efficient 3d model fitting using lifted
  optimization (2020)

\bibitem{kinect}
Shotton, J., Fitzgibbon, A., Cook, M., Sharp, T., Finocchio, M., Moore, R.,
  Kipman, A., Blake, A.: Real-time human pose recognition in parts from single
  depth images. In: CVPR (2011)

\bibitem{sitzmann2020implicit}
Sitzmann, V., Martel, J.N.P., Bergman, A.W., Lindell, D.B., Wetzstein, G.:
  Implicit neural representations with periodic activation functions (2020)

\bibitem{regcourse_sgp18}
Tagliasacchi, A., Bouaziz, S.: Dynamic 2d/3d registration. Proc. Symposium on
  Geometry Processing (Technical Course Notes)  (2018)

\bibitem{htrack}
Tagliasacchi, A., Schr{\"o}der, M., Tkach, A., Bouaziz, S., Botsch, M., Pauly,
  M.: Robust articulated-icp for real-time hand tracking. In: SGP (2015)

\bibitem{tancik2020fourierfeat}
Tancik, M., Srinivasan, P.P., Mildenhall, B., Fridovich-Keil, S., Raghavan, N.,
  Singhal, U., Ramamoorthi, R., Barron, J.T., Ng, R.: Fourier features let
  networks learn high frequency functions in low dimensional domains. arXiv
  preprint arXiv:2006.10739  (2020)

\bibitem{fscore}
Tatarchenko, M., Richter, S.R., Ranftl, R., Li, Z., Koltun, V., Brox, T.: What
  do single-view 3d reconstruction networks learn? In: CVPR (2019)

\bibitem{taylor2017articulated}
Taylor, J., Tankovich, V., Tang, D., Keskin, C., Kim, D., Davidson, P., Kowdle,
  A., Izadi, S.: Articulated distance fields for ultra-fast tracking of hands
  interacting. ACM Transactions on Graphics (TOG)  (2017)

\bibitem{thiery2016animated}
Thiery, J.M., Guy, {\'E}., Boubekeur, T., Eisemann, E.: Animated mesh
  approximation with sphere-meshes. ACM Transactions on Graphics (TOG)
  \textbf{35}(3),  1--13 (2016)

\bibitem{tkach2017online}
Tkach, A., Tagliasacchi, A., Remelli, E., Pauly, M., Fitzgibbon, A.: Online
  generative model personalization for hand tracking. ACM Transaction on
  Graphics (Proc. SIGGRAPH Asia)  (2017)

\bibitem{tung2017self}
Tung, H.Y., Tung, H.W., Yumer, E., Fragkiadaki, K.: Self-supervised learning of
  motion capture. In: Advances in Neural Information Processing Systems. pp.
  5236--5246 (2017)

\bibitem{tfgraphics}
Valentin, J., Keskin, C., Pidlypenskyi, P., Makadia, A., Sud, A., Bouaziz, S.:
  Tensorflow graphics: Computer graphics meets deep learning (2019)

\bibitem{nocs}
Wang, H., Sridhar, S., Huang, J., Valentin, J., Song, S., Guibas, L.J.:
  Normalized object coordinate space for category-level 6d object pose and size
  estimation. In: CVPR (2019)

\bibitem{disn}
Xu, Q., Wang, W., Ceylan, D., Mech, R., Neumann, U.: {DISN}: Deep implicit
  surface network for high-quality single-view 3d reconstruction. In: NeurIPS
  (2019)

\bibitem{rotations}
Zhou, Y., Barnes, C., Lu, J., Yang, J., Li, H.: On the continuity of rotation
  representations in neural networks. In: CVPR (2019)

\end{thebibliography}
\clearpage
\section{Acknowledgments}
We would like to acknowledge Hugues Hoppe, Paul Lalonde, Erwin Coumans, Angjoo Kanazawa, Alec Jacobson, David Levine, and Christopher Batty for the insightful discussions.
Gerard Pons-Moll is funded by the Deutsche Forschungsgemeinschaft (DFG,
German Research Foundation) - 409792180.
Andrea Tagliasacchi is funded by NSERC Discovery grant RGPIN-2016-05786, NSERC Collaborative Research and Development grant CRDPJ 537560-18, and NSERC Research Tool Instruments RTI-16-2018.

\section{Supplementary material}

\subsection{Additional discussion on tracking techniques}
\label{sec:additional}
In dense real-time tracking applications~\cite{schmidt2014dart,hadjust,taylor2017articulated,thiery2016animated}
there are examples of articulated models that use hybrid representations.
Similarly to our work, they also provide constant-time occupancy queries without the need for acceleration data structures.
However, these shape models consist of \textit{simple rigid primitives}, or use an \textit{artist-designed template}.
In contrast, we learn a deformable shape model from data, and allow users to query the pose-corrected occupancy anywhere in space.
In more detail:
\begin{itemize}
\item Schmidt et al.~\cite{schmidt2014dart}: relies on a discretization of the SDF on grids and/or primitives to answer distance queries,  only models piecewise \textit{rigid} parts (compare our rigid (R) variant), and requires the construction of the parts~SDF before tracking -- a process that relies on user interaction.
\item Tkach et al.~\cite{tkach2017online}: the quality of approximation produced by our method is vastly superior to that achieved by sphere-meshes.
Further, similarly to~\cite{schmidt2014dart}, the template is specified manually.
\item Taylor et al.~\cite{taylor2017articulated}: similarly to~\cite{schmidt2014dart}, the tracking template is specified manually, and the technique requires a \textit{mixture} of mesh-based closest point queries and implicit function queries.
\item Thiery et al.~\cite{thiery2016animated}: 
the approximation quality argument is analogous to that of~\cite{tkach2017online}, while to construct such models one need a consistently meshed motion sequence -- an extremely stringent requirement in practice.
\end{itemize}

\subsection{Ablation studies}
\label{sec:ablation}
Please see the animated results of \textit{reconstruction} and \textit{tracking} in the \textbf{supplementary video}. Please see the details of the dataset in the \textbf{supplementary data split files}.
Note that, except \Figure{ablation_pose} where we use AMASS/Transitions due to its diversity of poses,
we adopt AMASS/DFaust for all the other studies.
Also note that due to computational limitations, we evaluate on \textit{one} motion sequence only in~\Figure{ablation_modelsize}.
We select a sequence that has a median reconstruction performance as a representative example.


\begin{figure*}[h!]
\centering
\begin{minipage}{.45\columnwidth}
\centering
\scriptsize
\begin{tabular}{c|ccc}
\toprule
$\mathcal{L}_\text{occupancy}$ & mIoU$\uparrow$ & Chamfer~L1$\downarrow$ & F\%$\uparrow$ \\ 
\midrule
Cross-Entropy & .959 & .00006 & 98.01 \\
L2            & \textbf{.959} & \textbf{.00004} & \textbf{98.54} \\
\bottomrule
\end{tabular}
\vspace{-1em}
\captionof{table}{$\mathcal{L}_\text{occupancy}$}
\label{tbl:crossentropy}
\end{minipage}
\hspace{.1in}
\begin{minipage}{.45\columnwidth}
\centering
\scriptsize
\begin{tabular}{c|ccc}
\toprule
$\loss_\text{weights}$ & mIoU$\uparrow$ & Chamfer~L1$\downarrow$ & F\%$\uparrow$ \\ 
\midrule
\xmark & .845 & .00351 & 76.64 \\
\cmark & \textbf{.959} & \textbf{.00004} & \textbf{98.54} \\
\bottomrule
\end{tabular}
\vspace{-1em}
\captionof{table}{$\mathcal{L}_\text{weights}$}
\label{tbl:auxiliary}
\end{minipage}
\vspace{-0.4em}
\captionof{figure}{
Ablation study of the loss used for fitting the occupancy function (L2 vs. binary cross-entropy),
and the ablation study of the impact of the skinning weight loss in Eq.~\eqref{eq:skinningw-loss} on the right.}
\label{fig:losses}
\end{figure*}
\paragraph{Losses ablation -- \Figure{losses}}
One can view $\object(\x \given \pose)$ as a binary classifier that aims to separate the interior of the shape from its exterior.
Accordingly, one can use a binary cross-entropy loss for optimization, but our experiments suggest that an L2 loss perform
slightly better. Hence, we employ the L2 loss for all of our experiments; see~\Table{crossentropy}.
We also validate the importance of the skinning weights loss in~\Table{auxiliary} and observe a big improvement when
$\mathcal{L}_\text{weights}$ is included.

\begin{figure*}[h!]
\centering
\begin{minipage}{.32\columnwidth}
\centering
\scriptsize
%
\begin{tabular}{c|ccc}
\toprule
Model & mIoU$\uparrow$ & Chamfer~L1$\downarrow$ & F\%$\uparrow$ \\ 
\midrule
$R$                    & .933 & .00021 & 94.13 \\
$D{\setminus}\Pi$ & .926 & .00023 & 92.23 \\
$D$ & \textbf{.959}    & \textbf{.00004} & \textbf{98.54} \\
\bottomrule
\end{tabular}
%
\vspace{-1em}
\captionof{table}{Projection $\Pi$}
\label{tbl:projection}
\end{minipage}
\hspace{.5in}
\begin{minipage}{.52\columnwidth}
\centering
\scriptsize
%
\begin{tabular}{c|ccccc}
\toprule
$D$                                       & 1      & 2      & 4      & 8      & 16      \\ 
\midrule
mIoU$\uparrow$                          & .955   & .957   & \textbf{.959}   & .958   & .957    \\
Chamfer~L1$\downarrow$                  & .00130 & .00004 & .00004 & .00199 & .00004  \\
F\%$\uparrow$                           & 98.00  & 98.38  & \textbf{98.54}  & 98.09  & 97.85   \\
\bottomrule
\end{tabular}
%
\vspace{-0.4em}
\captionof{table}{Projection size $D$}
\label{tbl:projectionsize}
\end{minipage}
\vspace{-1em}
\captionof{figure}{Ablation of our (per-part) linear subspace projection.}
\label{fig:projection}
\end{figure*}
\paragraph{Linear subspace projection $\Pi$ -- \Figure{projection}}
Note that the rigid model~(R) actually \textit{outperforms} the deformable model~(D) if one \textit{removes} the learnt linear
dimensionality reduction ($D{\setminus}\Pi$); see~\Table{projection}.
This is a result only observed on the \textit{test} set, while on the training set $D{\setminus}\Pi$ performs comparably.
In other words, $\Pi$ helps our model to achieve better \textit{generalization} by enforcing a sparse representation of pose.
In~\Table{projectionsize}, we report the results of an ablation study on the dimensionality of the projection, which was the basis for the selection of~$D{=}4$.

\begin{figure*}[h!]
\centering
\begin{minipage}{.28\columnwidth}
\centering
\scriptsize
\begin{tabular}{c|c|ccc}
\toprule
 & $\pose$ & mIoU$\uparrow$ & ChamferL1$\downarrow$ & F\%$\uparrow$ \\ 
\midrule
D & $\{\T_b^{-1}\}$ & .962            & .00003                                  & 99.22           \\ 
D & $\{\T_b^{-1} \point\}$ & .959            & .00003                                  & 98.86           \\
D & $\{\T_b^{-1} \mathbf{t}_0\}$ & \textbf{.965}   & \textbf{.00002}                                  & \textbf{99.42}  \\ 
\bottomrule
\end{tabular}
\vspace{-1em}
\captionof{table}{$\pose$ for $D$.}
\label{tbl:pose_deformable}
\end{minipage}
\hspace{.4in}
\begin{minipage}{.6\columnwidth}
\centering
\scriptsize
\begin{tabular}{c|c|ccc}
\toprule
 & MLP input & mIoU$\uparrow$ & ChamferL1$\downarrow$ & F\%$\uparrow$ \\ 
\midrule
U & $[\point, \{\T_b^{-1}\mathbf{t}_0\}]$  & .520            & .001057                                 & 26.83           \\
U & $[\{\T_b^{-1}\point\}]$ & .865            & .00019                                  & 86.61           \\
D & $[\{\T_b^{-1}\point\}, \{\T_b^{-1}\mathbf{t}_0\}]$ & \textbf{.965}   & \textbf{.00002}                         & \textbf{99.42}  \\
\bottomrule
\end{tabular}

\vspace{-1em}
\captionof{table}{$[x,\pose$] for $U$ model.}
\label{tbl:pose_unstructured}
\end{minipage}
\vspace{-1em}
\captionof{figure}{Ablations of pose representations.}
\label{fig:ablation_pose}
\end{figure*}
\paragraph{Analysis of pose representations -- \Figure{ablation_pose}}
In \Table{pose_deformable}, we ablate several representations for the pose $\pose$ used by the deformable model.
We start by just using the \textit{collection} of homogeneous transformations $\{\T_b^{-1}\}$.
Note that the query point encoded in various coordinate frames is also an effective pose
representation~$\{\T_b^{-1} \point\}$, which has a much lower dimensionality.
Finally, we notice that rather than using the query point, one can pick a fixed point to represent pose.
While any fixed point 
can be used, we select the origin of the model $\mathbf{t}_0$ for simplicity, resulting in $\{\T_b^{-1} \mathbf{t}_0\}$.
The resulting representation is 
\textit{compact} and \textit{effective}.
\Table{pose_unstructured} shows a similar analysis for the unstructured model (metrics for D provided for reference).
Note how the performance of U can be significantly improved by providing the network with the encoding of the query point
$\point$ in various coordinate frames -- that is, the network is no longer required to ``learn'' the concept of changes of coordinates.

\begin{figure*}[h!]
\centering
\begin{minipage}{.4\columnwidth}
\centering
\scriptsize
\begin{tabular}{c|ccccc}
\toprule
Model & $24{\times}\mathbf{8}$ & $24{\times}\mathbf{16}$ & $24{\times}\mathbf{24}$ & $24{\times}\mathbf{32}$ & $24{\times}\mathbf{40}$ \\ 
\midrule
U & .539 & .538 & .601 & .642 & .653 \\
R & .913 & .902 & .931 & .939 & .946 \\
D & \textbf{.917} & \textbf{.915} & \textbf{.946} & \textbf{.950} & \textbf{.952} \\
\bottomrule
\end{tabular}
\captionof{table}{IoU metric.}
\label{tbl:ablation_modelsize}
\end{minipage}
%
\hspace{.05\linewidth}
\begin{minipage}{.45\columnwidth}
\centering
\begin{overpic}
[width=\linewidth]
{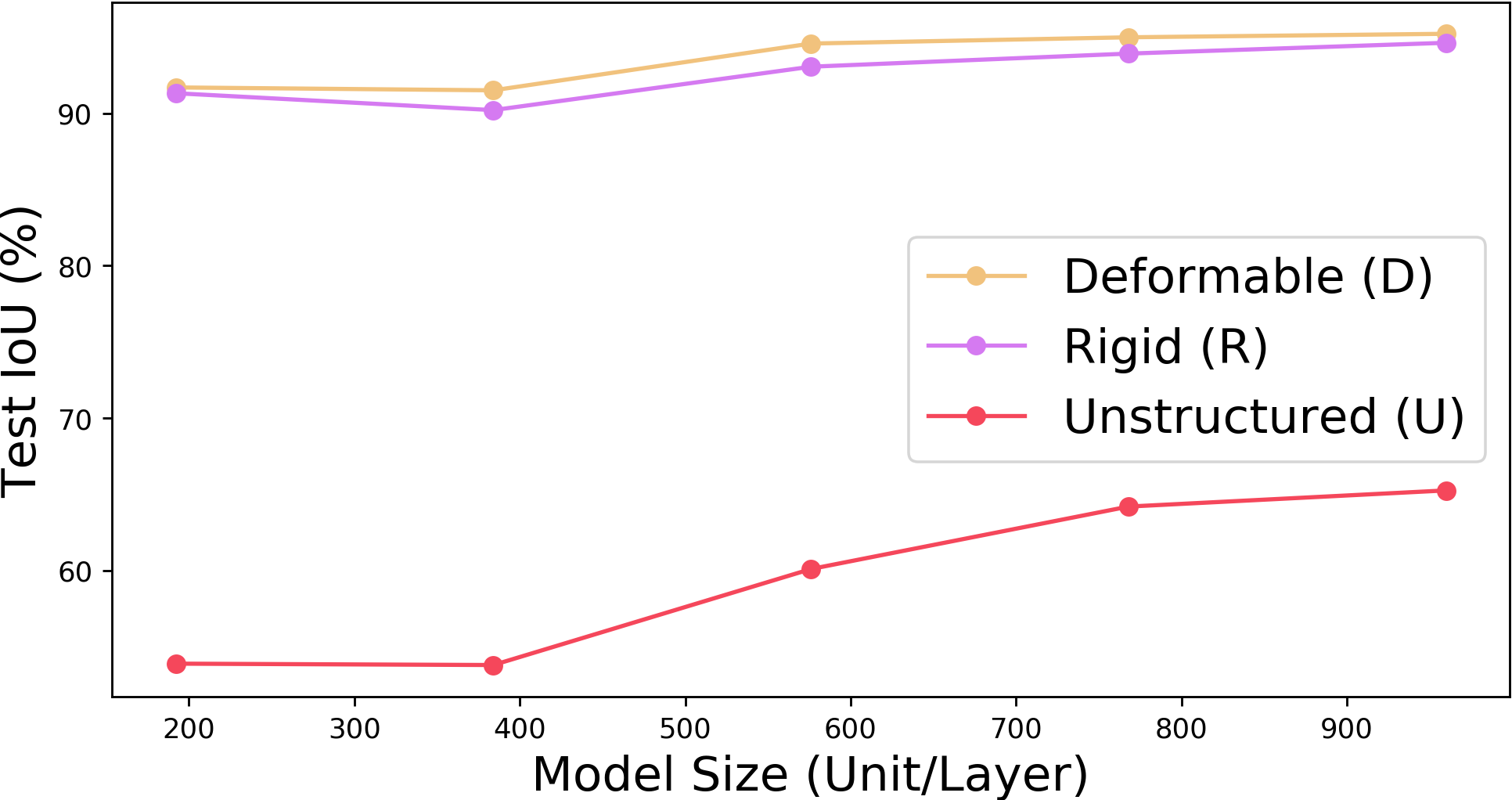}
\end{overpic}
\end{minipage}
\captionof{figure}{We evaluate the performance of the model as we increase the number of units used for each of the $24$ parts in the set $\{8,16,24,32,40\}$. The number of layers in each sub-network is held \textit{fixed} to $4$.}
\label{fig:ablation_modelsize}
\end{figure*}
\paragraph{Analysis of model size -- \Figure{ablation_modelsize}}
Both rigid (R) and deformable (D) models significantly outperform the results of the unstructured (U) model, as we increase the neural network's layer size and approach the network capacity employed by~\cite{imnet,deepsdf,occnet}.

\clearpage

\begin{figure*}[t]
\centering
\begin{minipage}{.49\columnwidth}
\centering
\resizebox{\linewidth}{!}{
\begin{tabular}{ccc|ccc}
\toprule
$p(\mathcal{O} \given \pose)$ & $p(\pose)$ & $\circledast$ & mIoU$\uparrow$ & Chamfer~L1$\downarrow$ & F\%$\uparrow$ \\ 
\midrule
D & \xmark & \xmark & .952 & .00005 & 97.24 \\
D & \xmark & \cmark & .948 & .00005 & 97.50 \\
D & \cmark & \xmark & .965 & .00004 & 98.79 \\
D & \cmark & \cmark & \textbf{.968} & \textbf{.00004} & \textbf{99.08} \\
\bottomrule
\end{tabular}
} 
\vspace{-1em}
\captionof{table}{DFaust ``easy'' (00-01)}
\end{minipage}
\begin{minipage}{.49\columnwidth}
\centering
\resizebox{\linewidth}{!}{
\begin{tabular}{ccc|ccc}
\toprule
$p(\mathcal{O} \given \pose)$ & $p(\pose)$ & $\circledast$ & mIoU$\uparrow$ & Chamfer~L1$\downarrow$ & F\%$\uparrow$ \\ 
\midrule
D & \xmark & \xmark & .546 & .01430 & 44.31 \\
D & \xmark & \cmark & .891 & .00032 & 86.15 \\
D & \cmark & \xmark & .862 & .00258 & 79.05 \\
D & \cmark & \cmark & \textbf{.948} & \textbf{.00006} & \textbf{96.48} \\
\bottomrule
\end{tabular}
} 
\vspace{-1em}
\captionof{table}{DFaust ``hard'' (02-09)}
\end{minipage}
\vspace{-0.5em}
\captionof{figure}{Ablations for the tracking application}
\label{fig:ablation_tracking}
\end{figure*}
\paragraph{Tracking ablations -- \Figure{ablation_tracking}}
In the tracking application, we ablate with respect to the pose prior ($p(\pose)$) and the use of random perturbations to approximate the distance function via convolution~($\circledast$).
First, note that the best results are achieved when both of these components are enabled, across \textit{all} metrics.
We compare the performance of our models on easy vs.~hard sequences. Hard sequences more clearly illustrate the advantages of the algorithms proposed.
We validate the usefulness of the pose prior in avoiding tracking failure~(e.g. $IoU: 44.31\% \rightarrow 86.15\%$).
The use of random perturbations allow the optimization to converge more precisely~(Chamfer: $.00258 \rightarrow .00006$).

\begin{figure*}[h!]
\centering
\begin{overpic}
[width=.7\linewidth]
{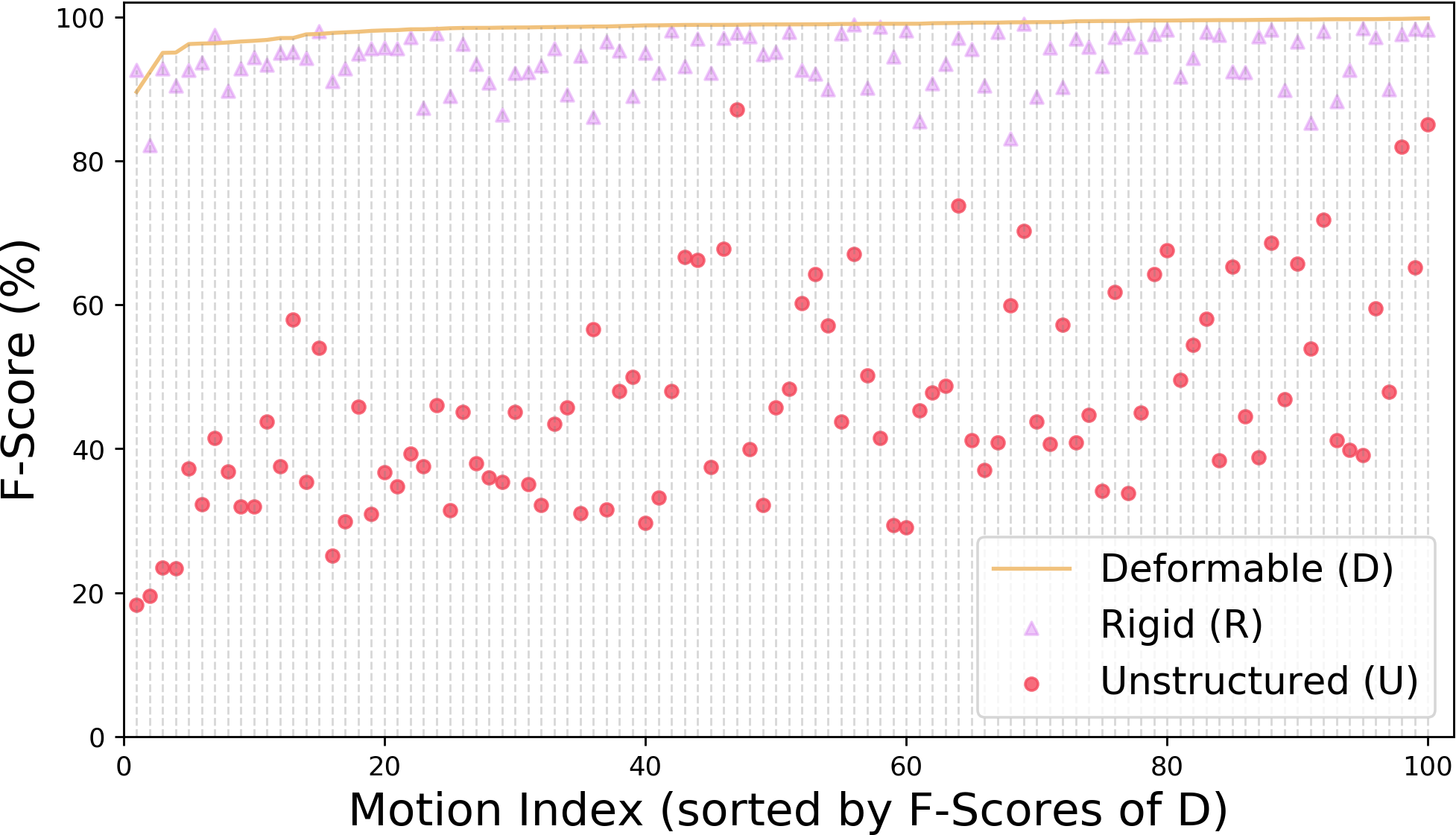}
\end{overpic}
\vspace{-.5em}
\caption{
Distribution of F-Score across the AMASS/DFaust dataset.
}
\label{fig:ablation_distribution}
\end{figure*}
\paragraph{Metrics distribution on AMASS/DFaust}
Rather than reporting aggregated statistics, we visualize the IoU errors of all of the $100$ DFaust experiments, and sort them by the performance achieved by the deformable model~(D).
Note how the deformable model achieves consistent performance across the dataset. There are \textit{only two sequences} where the rigid model performs better than the deformable model.

\clearpage
\end{document}